\documentclass[letterpaper]{article} 
\usepackage{aaai25}  
\usepackage{times}  
\usepackage{helvet}  
\usepackage{courier}  
\usepackage[hyphens]{url}  
\usepackage{graphicx} 
\usepackage{natbib}  
\usepackage{caption} 
\usepackage{algorithm}
\usepackage{algorithmic}
\usepackage{xspace}

\usepackage{newfloat}
\usepackage{listings}
\DeclareCaptionStyle{ruled}{labelfont=normalfont,labelsep=colon,strut=off} 
\floatstyle{ruled}
\newfloat{listing}{tb}{lst}{}
\floatname{listing}{Listing}
\usepackage[colorlinks=true,linkcolor=black,urlcolor=blue,citecolor=black]{hyperref}

\usepackage{graphicx}
\usepackage{amsmath}
\usepackage{amssymb}
\usepackage{booktabs}
\usepackage{comment}
\usepackage{bm}
\usepackage{url}
\newcommand{\link}[2]{\href{#2}{#1}}

\newcommand{\datasetname}{SBS Figures\xspace}
\title{SBS Figures: Pre-training Figure QA from Stage-by-Stage Synthesized Images}
\author{
    Risa Shinoda\textsuperscript{\rm 1}\textsuperscript{\rm 2},
    Kuniaki Saito\textsuperscript{\rm 2},
    Shohei Tanaka\textsuperscript{\rm 2},
    Tosho Hirasawa\textsuperscript{\rm 2},
    Yoshitaka Ushiku\textsuperscript{\rm 2}
}
\affiliations {
    \textsuperscript{\rm 1}Kyoto University, Japan\\
    \textsuperscript{\rm 2}OMRON SINIC X Corp., Japan\\
    \{risa.shinoda, kuniaki.saito, shohei.tanaka, tosho.hirasawa, yoshitaka.ushiku\}@sinicx.com
}

\usepackage{bibentry}
\usepackage{etoolbox}
\makeatletter
\patchcmd{\NAT@citex}
  {\@citea\NAT@hyper@{\NAT@nmfmt{\NAT@nm}\hyper@natlinkbreak{\NAT@date}}}
  {\@citea\NAT@nmfmt{\NAT@nm}\NAT@date}
  {}{}
\makeatother

\begin{document}
\nocopyright 
\maketitle

\begin{abstract}
Building a large-scale figure QA dataset requires a considerable amount of work, from gathering and selecting figures to extracting attributes like text, numbers, and colors, and generating QAs. Although recent developments in LLMs have led to efforts to synthesize figures, most of these focus primarily on QA generation. 
Additionally, creating figures directly using LLMs often encounters issues such as code errors, similar-looking figures, and repetitive content in figures.
To address this issue, we present \datasetname~(Stage-by-Stage Synthetic Figures), a dataset for pre-training figure QA. Our proposed pipeline enables the creation of chart figures with complete annotations of the visualized data and dense QA annotations without any manual annotation process.
Our stage-by-stage pipeline makes it possible to create diverse topic and appearance figures efficiently while minimizing code errors.
Our \datasetname~demonstrate a strong pre-training effect, making it possible to achieve efficient training with a limited amount of real-world chart data starting from our pre-trained weights.
Our code is available at \link{https://github.com/omron-sinicx/SBSFigures}{https://github.com/omron-sinicx/SBSFigures}.

\end{abstract}

\section{Introduction}
\label{sec:intro}

Building models that understand figures is essential for automating document understanding, given that numerous documents incorporate figures for data visualization. 
Understanding figures necessitates two key abilities of models: (i) precise interpretation of visualized information, encompassing numerical data, labels, and plot positions, and (ii) reasoning to return accurate responses based on the visualized information and user queries. 
A densely annotated dataset with many graph figures is imperative to train such models effectively. Given the requirements above, these annotations need to (i) offer exhaustive details about each figure, e.g., the values and labels of plots along with the graph title, and (ii) include question-answer pairs to build a well-performing QA model~\cite{chartqa}.

 \begin{figure}[t]
  \centering
  \includegraphics[width=\columnwidth]{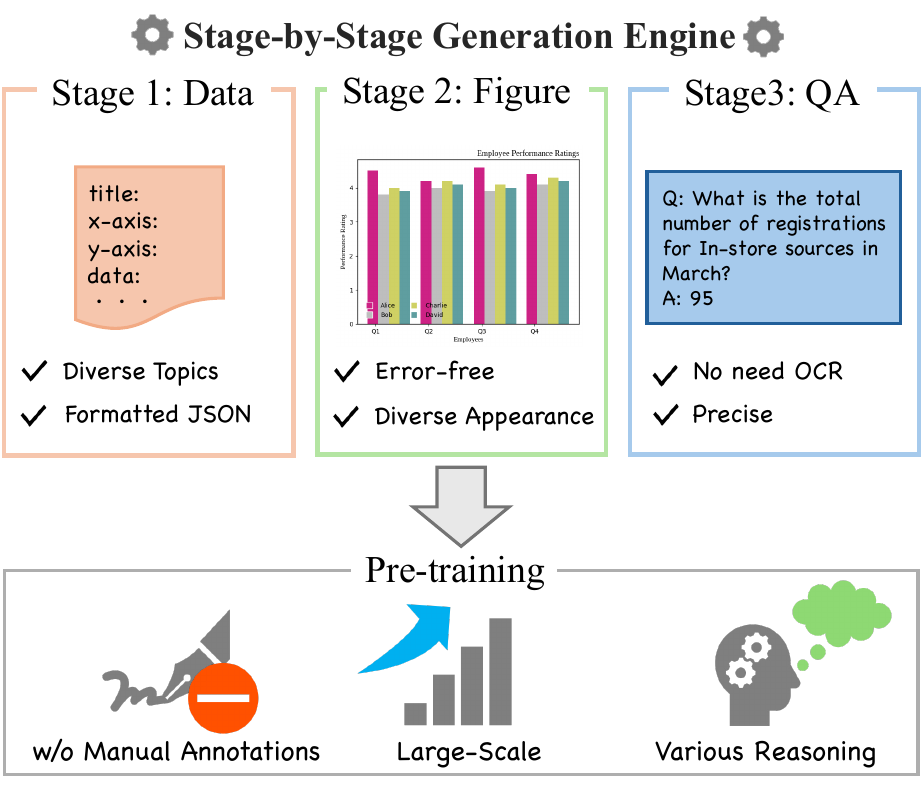}
\caption{\textbf{SBS Figures~(Stage-by-Stage Synthetic Figures).} 
We create \datasetname, a dataset for pre-training figure QA.
Our stage-by-stage synthetic dataset creation enables a strong pre-training effect for real-world chart data.}

  \label{fig:theser}
  \vspace{-9pt}
\end{figure}

However, collecting such a dataset is not easy. First, collecting figures itself needs careful selection and parsing of diverse documents or websites~\cite{Siegel2016FigureSeerPR}, and its cost will increase if we care about their copyright.
Second, exhaustive labeling demands a lot of effort for annotators as they must first comprehend the visualization and then translate it into texts~\cite{hoque2022chart}. 
To reduce the extensive annotation effort, prior approaches have utilized template-based QA augmentation or LLMs~\cite{figureqa, plotqa, reasoning}. While this reduces the cost of QA generation, the number of figures remains limited. Recently, LLM-based figure generation has been proposed (e.g., ChartLlama~\cite{chartllama}) 
, where an LLM generates both a visualization target and code to render the target for each figure.
However, this pipeline is inefficient in terms of the number of queries required, as each figure demands a separate LLM query. Moreover, the synthesized code often contains errors without refinement.

To address this challenge, we propose a novel stage-by-stage pipeline for generating both figures and their annotations, designed to progressively transform the seed to synthesize a figure. Specifically, our pipeline divides the figure generation process into three modules, visualization target data generation, figure rendering via Python code, and QA pair generation, with each module progressively transforming seed data. This stage-by-stage approach offers three key advantages over generating figures all at once: 
(i) the generated seed data, such as data point to be visualized, and figure rendering code, can be stored and reused, reducing the cost of querying LLMs multiple times,
(ii) the figure’s appearance can be easily diversified, as components like chart type, data content, and font can be controlled at each stage, and (iii) figure rendering is more reliable because, rather than having the LLM generate rendering code for each data point, we create the figure-rendering code separately for each figure type, using a same data structure.

Using our scalable dataset generation pipeline, we introduce a new dataset called \datasetname~(Stage-By-Stage Synthetic Figures), which comprises 1 million figure images, each paired with annotations of the accurate visualized data and QA pairs. 
Our model, pre-trained on \datasetname, demonstrates strong performance on real-world figure QA datasets, enabling efficient learning on real-world charts. 
Our \datasetname pre-training shows generality to models and fine-tuning dataset.
Beyond the pipeline and dataset, we also investigate key factors influencing figure QA pre-training. 
Specifically, we explore the impact of various dataset components on pre-training with synthetic datasets, including figure appearance, QA quality, and task prompts, which have not been fully explored in previous studies. 
We will make the entire pipeline, including the dataset, code, prompts for LLM, and models, publicly available, allowing future research to use our \datasetname for more efficient training.

In summary, our main contributions are as follows:

\begin{itemize}
    \item We propose a new pipeline that enables the efficient creation of diverse topics, visually distinct figures, and dense QAs while minimizing code errors.
    \item Our proposed dataset, entirely synthesized, demonstrates a strong pre-training effect on real-world chart data. This dataset enables effective pre-training even with a limited amount of real-world chart data.
    \item We will make the entire pipeline, including the dataset, code, prompts, and models, publicly available.
\end{itemize}

\begin{figure*}[t]
  \centering
  \includegraphics[height= 9cm]{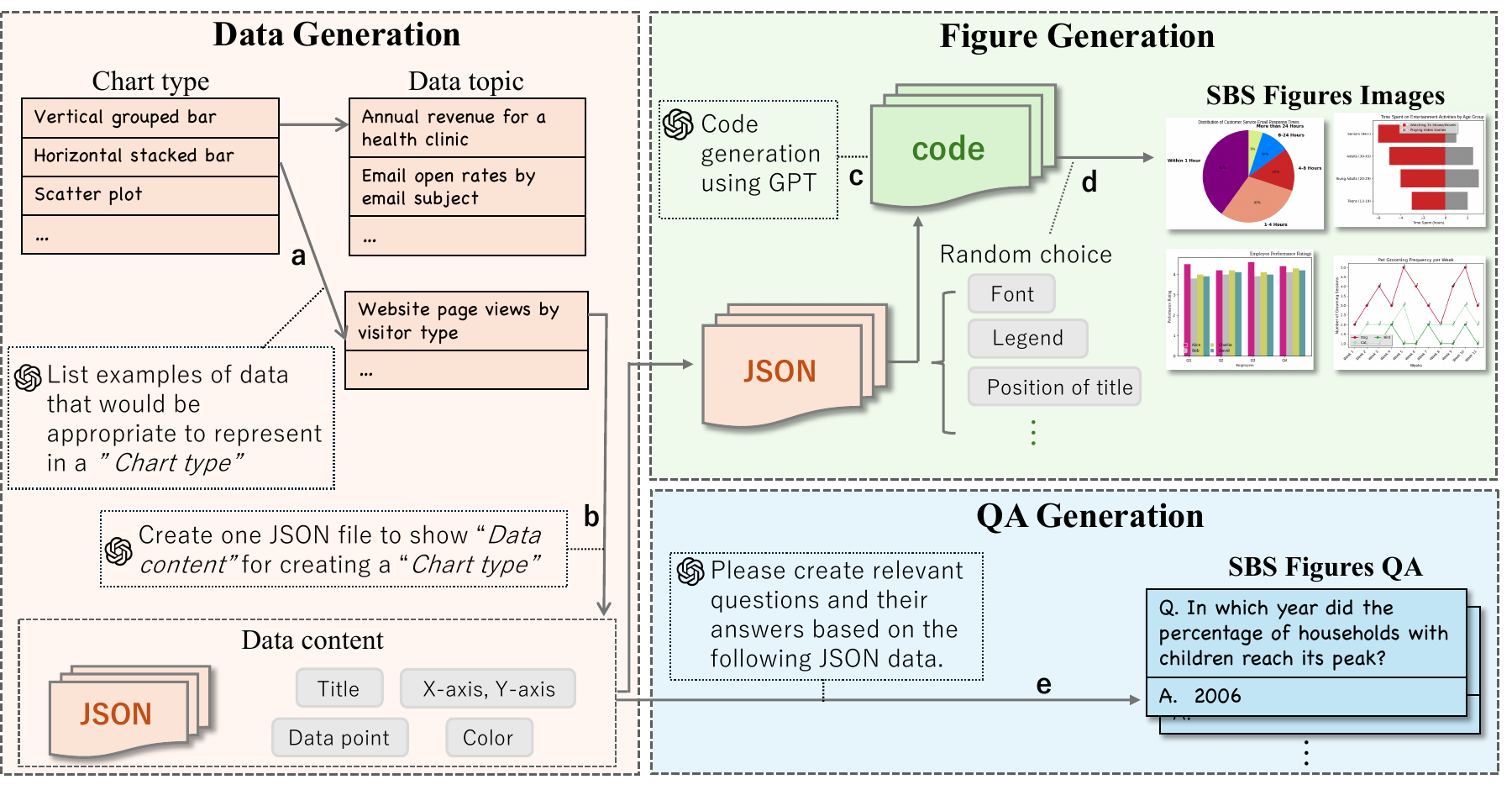}
\caption{\textbf{Generation pipeline of \datasetname.} \datasetname~was created using a fully synthetic method. First, we generate the visualization data, represented in JSON format, containing complete numbers, text, and colors. Next, we produce figure images from this data using pre-defined, error-free Python scripts. Finally, we generate dense and accurate QA pairs from visualization data without the need for OCR.}

  \label{fig:flow}
  \vspace{-9pt}
\end{figure*}

\section{Related Work}
\noindent\textbf{Synthetic figure generation}
To overcome the limited number of annotated real-world figures, some existing datasets create figures in two main ways. 
The first way is gathering real-world data and converting them into figures: datasets such as LEAF-QA~\cite{leafqa}, LEAFQA++~\cite{leafqaplus}, and PlotQA~\cite{plotqa} have been introduced.
However, collecting real-world data suitable for synthetic figure generation remains a costly process, therefore, the number of figure images is still limited (e.g., LEFQA / LEAFQA++: 250k, PlotQA: 224k).
To address the costly process of data curation, previous works have attempted to synthesize data for visualizing figures, leading to the development of datasets like FigureQA~\cite{figureqa} and DVQA~\cite{dvqa}. FigureQA uses fixed labels based on bar color, while DVQA randomly selects words from vocabularies in its two dataset splits. However, this creates a big gap from real-world figure data, as the data itself lacks meaningful content due to the use of random words and fixed vocabularies.

Recently, ChartLlama~\cite{chartllama} 
\footnote{We don't compare with this because it is concurrent and the dataset is not openly available yet.\label{footnote1}}  
created figures by providing data topics and trends to GPT-4, resulting in the generation of 160k synthetic figures. 
Differently from this work, we generate figures synthetically from data topics, enabling us to create as many figures as needed. 
Additionally, we use pre-defined Python code to ensure error-free and efficient figure generation, enabling the large-scale production of synthetic figures.

\noindent\textbf{Synthetic figure QA}
Generating QA is the next step from generating figures. 
While precise human annotation~\cite{chartqa} enables complex task evaluation, 
creating massive QA datasets is also important for developing strong models. 
Initially, FigureQA~\cite{figureqa} created template-based Yes/No questions. 
To more complicated QA pairs, fixed vocabulary template-based datasets have been developed~\cite{leafqa,leafqaplus,dvqa}. LEAFQA~
\footref{footnote1}~\cite{leafqa} and LEAFQA++~
\footref{footnote1}~\cite{leafqaplus} use 35 and 75 templates each, and DVQA uses 26 templates.

While template-based QA reduces the need for extensive annotation in QA generation, more diverse methods for generating QA pairs have been explored.
PlotQA~\cite{plotqa} synthesizes open-vocabulary QA using 74 annotated template-based question generations. 
For more diverse QA generation without templates,
ChartQA-Synth
~\footref{footnote1} was recently generated, adding 544k QA pairs to the ChartQA dataset~\cite{chartqa} using LLMs~\cite{reasoning}.
Similarly,~\citeauthor{Li_2024_CVPR}
created synthetic QA pairs for existing 312k images from the ChartQA dataset~\cite{chartqa} with LLMs~\cite{Li_2024_CVPR}.
In contrast, our approach involves using LLMs to generate both the figures and the QA pairs from scratch, allowing us to greatly expand the dataset with a large number of figures and corresponding presence data and QA pairs.

\noindent\textbf{Models for figure understanding}
Multi-modal models that combine visual and text modalities have been developed~\cite{t5,vlt5,donut, pix2struct}.
For robust visual language models (VLMs), pre-training on synthetic datasets is an effective strategy. 
For instance, Pix2Struct~\cite{pix2struct} trained ViT-based models by converting images to HTML representations, while Donut~\cite{donut} utilized synthetic datasets to extract and read textual information from images during pre-training.
These pre-training strategies enhance the model's performance.

Building a VLM for Figure QA is a well-researched area. These tasks involve both OCR and reasoning over data points, requiring specialized models. One effective way to build a strong Figure QA model is by training it with figure-related data.
UniChart~\cite{UniChart}, building on Donut~\cite{donut}, is fine-tuned using figure-specific datasets to improve performance. 
Similarly, Matcha~\cite{matcha} is a figure-focused model pre-trained using figure-related tasks and builds on Pix2Struct~\cite{pix2struct}. 
For the combination of LLM, Deplot~\cite{deplot} used a hybrid approach by converting figures into JSON-style data for further reasoning with large language models (LLMs).
Our proposed pipeline enhances this model training process by generating synthetic figures and their corresponding data points, providing tailored datasets for pre-training figure-specific VLM models.

\section{\datasetname~Dataset}

To overcome the limitations in existing synthetic figure QA datasets, we designed our dataset creation pipeline to meet four criteria: 
(1) The figures should cover a diverse range of topics and types, and exhibit a variety of visual styles.
(2) Efficiently create figures for generating a large-scale dataset with error-free code.
(3) The question-answer pairs must accurately reflect the information presented in the figures, and  
(4) All instances are copyright-free.
To fulfill criteria (1) and (2), our pipeline initially uses an LLM to generate the target data for visualization. Each piece of data is then converted into figures and associated question-answer pairs. By pre-defining the figure generation code and integrating randomness into each component, we achieve both efficiency and variety in the generation process.
In our work, we mainly use the GPT-3.5-turbo as our backbone LLM, ensuring its outputs are copyright-free to meet criterion (4).
The entire generation process is illustrated in Figure~\ref{fig:flow}. See the Figure~\ref{fig:prompt} for detailed prompts.

\subsection{Data Generation}
\label{sec:datagene}
\begin{figure*}[t]
  \centering
  \includegraphics[height= 6.5cm]{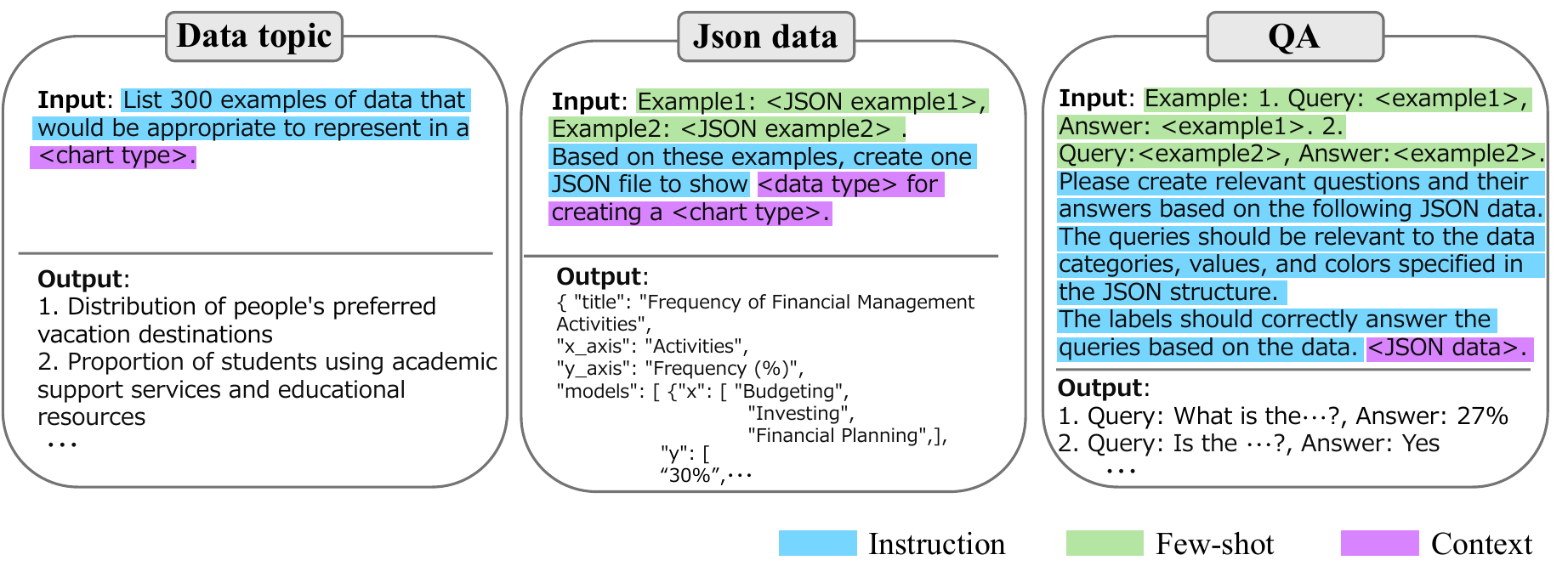}
\caption{\textbf{Prompt templates used in the generation pipeline of \datasetname.} We adopt few-shot prompting to ensure consistent formatting for both JSON data and QA generation. To improve efficiency, our pipeline includes code that repeatedly adjusts the context and prompts during the generation process.}
  \label{fig:prompt}
\end{figure*}

Here, we describe how to generate data points that will be converted to figures and QAs as shown in the left of Figure~\ref{fig:flow}. We propose a two-step data generation approach to enhance the diversity of the output. In the first step, a data topic is generated, followed by the data content in the second step. Since the data topic acts as a seed for generating the visualized data, this method allows for greater variability in the results.

\noindent{\textbf{Data topic}}
We generate various data topics to be visualized using an LLM. Specifically, we synthesize figure topics given a chart type from the pre-defined commonly used ten figure types: Diverging bar chart, Vertical / Horizontal bar chart, Vertical / Horizontal grouped bar chart, Vertical / Horizontal stacked bar chart, Line chart, Scatter plot, and Pie chart~(Figure~\ref{fig:flow} a).
To cover diverse topics, we generate topics and construct over 100k unique topics for each chart type. We query the prompt shown in Figure~\ref{fig:prompt} (data topic) multiple times and delete duplicates to ensure the variety of the data to be visualized.

\noindent{\textbf{Data content}}
We pre-defined the JSON format for each chart type.
Given the data topics, we ask LLM to create figure data to illustrate each topic~(Figure~\ref{fig:flow} b). 
We adopt JSON-style data representation as in previous works~\cite{deplot}, which can easily include a title, axis, and corresponding colors along with data points.
To ensure data-style consistency, we adopt few-shot prompting using example JSONs for each chart type. To enhance the diversity of data styles, we create around 10 examples per chart type by varying the number of data points and data trends, then randomly select them for few-shot prompting.
We finally obtain JSON files, which include the title, x, y-axis, data labels, data point numbers, and corresponding colors. 
This two-step data generation process ensures the diversity of topics in our dataset and consistency of the data formatting. 

\subsection{Figure generation}
\label{sec:figuregene}
We adopt a step-by-step data-to-figure rendering process because LLMs often generate code with errors when generating data and code simultaneously.

\noindent{\textbf{Code generation}}
Firstly, we create one chart generation code per chart type instead of having one code per JSON file. 
We also pre-defined the JSON format for each chart type in the data generation process, allowing us to create a generic figure generation code that can handle any data following the defined data structure. 
This approach avoids code errors and does not require LLM to generate code for each chart individually, making the process more efficient and suitable for large-scale figure generation.
For the creation of the pre-defined Python code~(Figure~\ref{fig:flow} c), we use GPT-4 due to its superior coding capabilities than GPT-3.

Then, we feed the JSON data to the code to generate the figures~(Figure~\ref{fig:flow} d). We randomized following components: 
\begin{itemize}
    \item  Fonts: We randomly select fonts used in figures from seven commonly used types.  We also use various font sizes.

    \item  Title: We randomly position the title at the center, middle, or right. Additionally, since real-world figures often lack titles, we also randomly create no-title figures.

    \item  Legend: We randomize the presence of legends if the chart type does not necessarily require legend information. We also randomly select the legend location from six positions.

    \item  Marker: We randomly choose the marker style from nine types.

    \item  Spine: We randomly select the presence of spines.

    \item  Numbers: We randomly select whether to show the represented numbers.
\end{itemize}
\noindent{\textbf{Convert data to the figure images}}
We randomize the chart properties in the generation code to create a variety of figures, e.g., title position, colors, fonts, marker style, the presence of gridlines, and whether to include numbers on the figures. 
This randomness results in around 2,000 unique combinations per chart type, greatly enhancing the variety in the appearance of the generated figures.

\subsection{QA generation}
We generate reliable QA pairs from JSON data used for figure creation.

\noindent{\textbf{QA creation by LLM}}
\begin{figure*}[t]
  \centering
  \includegraphics[width=\textwidth]{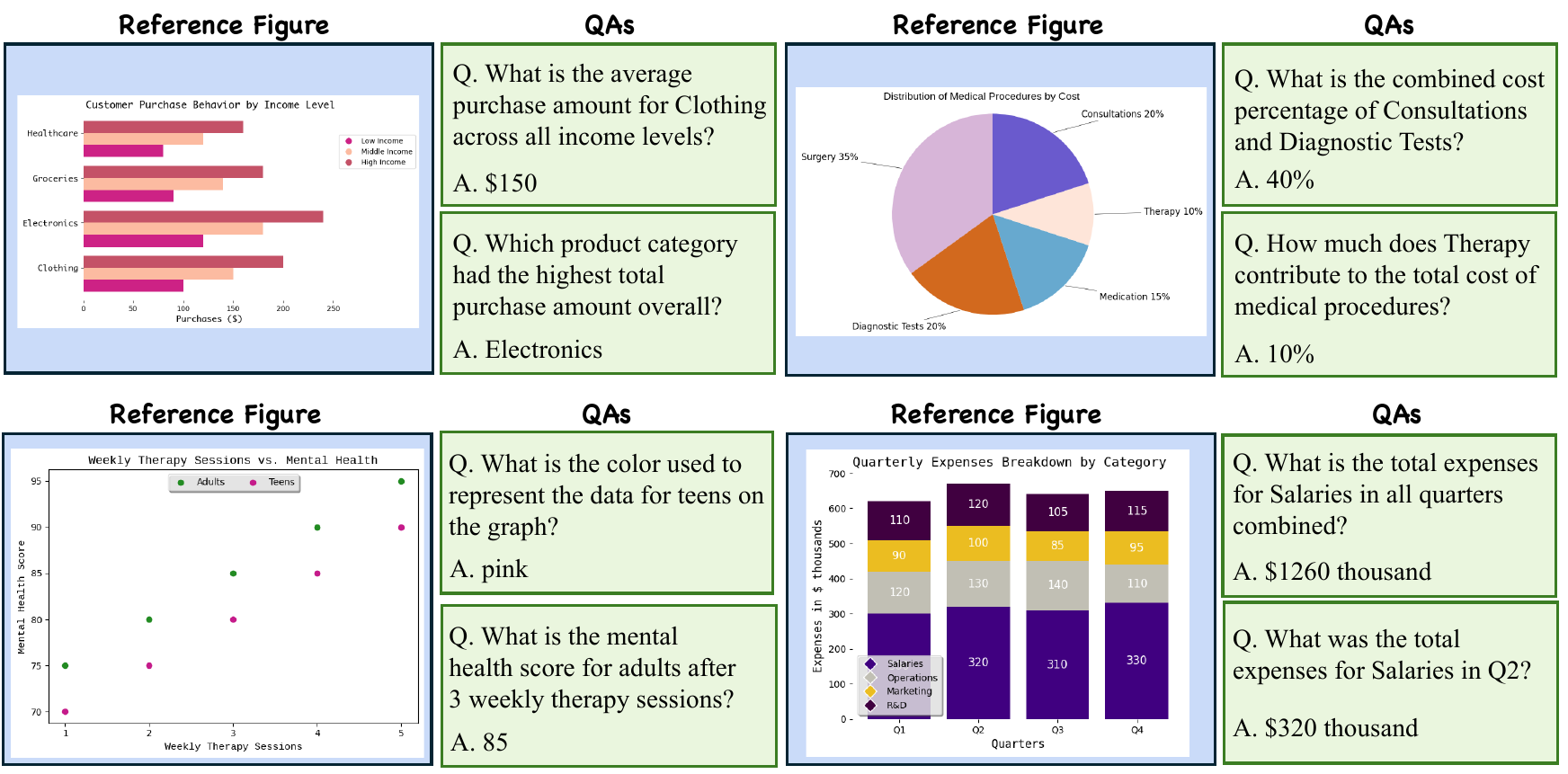}
\caption{\textbf{Example of \datasetname~QA pairs.} 
The figures show diverse visual variations, with each data content containing around 2,000 combinations of visual components. Additionally, our pipeline generates dense and precise QA pairs, requiring complex reasoning skills to address the questions.}
  \label{fig:ex_qa}
\end{figure*}
We provide generated JSON data from~\ref{sec:datagene} to LLM and ask to generate QA pairs based on the provided data~(Figure~\ref{fig:flow} e).
Since the JSON data includes the complete annotation for the data visualized in the figure, LLM can make QA pairs without the OCR process, making the QA numbers and calculations reliable.
Compared to human QA annotation from figures, we can easily access the data points, making it straightforward even for figures that do not explicitly display numerical values. We use few-shot prompting to format the QA pairs. See the detailed prompting in the Figure~\ref{fig:prompt} (QA). 
Here, we pre-define approximately 10 types of questions for each chart type and randomly include two examples as a context for each prompt. 
These questions are designed to test various reasoning skills, such as computation, data extraction, and color identification.
In this phase, LLM generates several QA pairs based on each JSON file. 

\subsection{Statistics}
\begin{figure}[t]
  \centering
  \includegraphics[width= 0.5\textwidth]{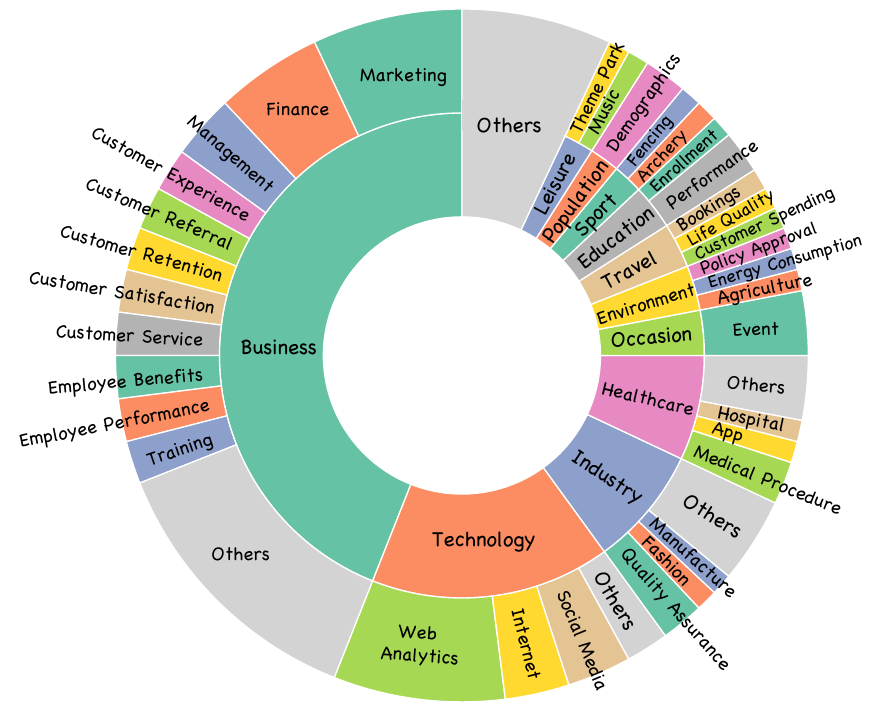}
\caption{\textbf{Theme distribution of \datasetname.} We randomly select 10 questions from each figure type and manually analyze the topic of the figure.}
  \label{fig:theme_stat}
\end{figure}
We finally obtained 1M images of \datasetname. \datasetname~
consists of 10 types of figures and includes 4.2M dense QA pairs with complete JSON format data, including title, axis, data, and colors.
Each type of figure is generated with around 2,000 combinations of appearance variations, which are defined within the Python code.
Our \datasetname~is free from copyright issues.
We illustrate the example images and QAs in Figure~\ref{fig:ex_qa}.

To show the distribution of our \datasetname themes, we randomly selected 10 figures from each type and manually categorized them using two hierarchical levels. As shown in Figure~\ref{fig:theme_stat}, \datasetname demonstrates a wide range of themes. The largest category is business, covering a wide range of topics such as marketing and finance. Although users can control the data topics in their prompts by specifying the topic or modifying the provided examples, our default prompt allows for the generation of figures on diverse topics.

\section{Experiments}
In this section, we conduct experiments to demonstrate the effectiveness of \datasetname. 
First, we conduct main experiments to show our \datasetname generality.
Then, we examine which aspects of the synthetic dataset are important for pre-training figure QA tasks in the investigation results section. 
Since \datasetname~can be created without manual annotations, we generated multiple variations of \datasetname~by changing different factors. We then conducted experiments to identify the key factors that affect figure QA pre-training.
\datasetname.

\subsection{Experimental Settings}

\noindent\textbf{Dataset and Evaluation}
Here, we describe the dataset details used in our experiments.

\begin{itemize}
\item \noindent{\textbf{ChartQA}}~\cite{chartqa} is a standard and challenging dataset for evaluating document understanding models. This dataset contains two splits of QA pairs: 1) human split (human), which is annotated by humans, and 2) augmented split (aug.), which is created by the T5~\cite{t5} model. The human split has 9.6k QA pairs, the augmented split has 23.1k pairs, and the total number of figure images is 21k.

\item \noindent{\textbf{PlotQA}}~\cite{plotqa} is a synthesized dataset created by converting real-world data into figures. QA pairs are generated using templates created by human annotators. There are two splits of the dataset: v1, which has around 8M QA pairs, and v2, which is an extended version with 20M QA pairs. Both splits share the same set of figure images, totaling 224k images.

\item \noindent{\textbf{FigureQA}}~\cite{figureqa} is a synthesized dataset created using fixed vocabularies representing 100 colors. The QAs are generated using 15 templates, and the answer type is limited to yes (1) or no (0) questions. The dataset contains 2.3M QA pairs and 180k figure images.

\item \noindent{\textbf{DVQA}}~\cite{dvqa} is a synthesized dataset created using 1K fixed nouns. The QAs are generated using 26 templates, and the answers come from closed vocabularies. The dataset contains 3.4M QA pairs and a total of 300k figure images.
\end{itemize}
We don't compare with LEAFQA~\cite{leafqa}, LEAFQA++~\cite{leafqaplus}, ChartQA-Synth~\cite{reasoning}, and ChartLlama~\cite{chartllama}, because they are not openly available yet.
We primarily evaluate on the ChartQA dataset~\cite{chartqa}.
Following previous works~\cite{UniChart, matcha}, we use exact match accuracy with numerical tolerance for 5\%.

\noindent\textbf{Model}
Our experiments are mainly based on the Donut~\cite{donut}. We also use Pix2Struct~\cite{pix2struct} to show our dataset generality.
\begin{itemize}
    \item Donut~\cite{donut}: an openly available OCR framework. The encoder is built on the Swin Transformer~\cite{swin} architecture, and the text decoder is BART~\cite{bart}. We initialized the model weight with the donut-base model. The model parameter count is 201M.
    \item Pix2Struct~\cite{pix2struct}: an image-encoder-text-decoder based
on ViT~\cite{ViT}. This model is pre-trained by parsing masked screenshots of web pages into HTML. We initialized the model weight with the pix2struct-textcaps-base model. The model parameter count is 282M.
\end{itemize}
For the Donut model implementations, we follow UniChart~\cite{UniChart} for hyperparameters. In the pre-training phase, we use a batch size of 80 and a learning rate of 1e-4. In the fine-tuning phase, we use a batch size of 24 and a learning rate of 5e-5. The input image resolution is set to $960 \times 960$. We apply a cosine scheduler with a warm-up step of 100. For the Pix2Struct model, the learning rate and batch size are the same as those of Donut. The only changes we make are adjusting the image resolution to $640 \times 640$ and decreasing the pre-training batch size to 40, due to the larger model size and higher computational resource requirements. For computational resources, we use A100 GPUs, with a minimum of 4 GPUs and a maximum of 8 GPUs, depending on the batch size.

Excluding the noted parts, we conduct pre-training for 3 epochs for each dataset and fine-tune the model for 20 epochs on the ChartQA~\cite{chartqa} dataset. For fine-tuning on the PlotQA~\cite{plotqa} and FigureQA~\cite{figureqa} datasets, we train for 1 epoch on each training split and test on a subset of 10k QA pairs from the test split.
\subsection{Main Results}
\label{subsec:main_results}
\noindent\textbf{Dataset Comparison} 
We evaluate the pre-training effectiveness of our \datasetname~compared to other synthetic figure QA datasets by conducting pre-training on each dataset and fine-tuning each model on ChartQA. \footnote{We don't compare with LeafQA, LeafQA++, and ChartLlama, because these datasets are not openly available yet.\label{footnote2}}  
Pre-training was conducted for 3 epochs, except for PlotQA, which was pre-trained for 1 epoch due to the difference in QA numbers. 
After pre-training of each dataset, all models were fine-tuned for 20 epochs for ChartQA.
Pre-training and fine-tuning were performed using the Donut~\cite{donut} model. 
In this context, "scratch" refers to fine-tuning on ChartQA dataset without any pre-training, starting from the vanilla Donut-base weights.
As shown in Table~\ref{tab:comparison-data}, only \datasetname~demonstrates the improvements by pre-training.
Other datasets, which contained only template-based QA, did not show a pretraining effect.  
PlotQA, which is based on real-world data, achieved superior results compared to the others. 
We further assess the contributions of the components of our \datasetname~by investigation results section.

\begin{table}[t]
  \centering
  \begin{tabular}{llll}\toprule[0.8pt]
    \textbf{Dataset}    & \textbf{human} & \textbf{aug.} & \textbf{avg}\\
    \midrule
    Scratch      &    31.28  & 77.76   &   54.42                \\
    FigureQA~\cite{figureqa}      &      13.44      &       9.36        & 11.40         \\
    DVQA~\cite{dvqa} & 26.88    &     72.16    &    49.52           \\
    PlotQA~\cite{plotqa}   &     30.56              & 74.00   &52.28 \\
    \datasetname(Ours)   &   \textbf{39.44}    & \textbf{82.24}   &  \textbf{60.84}   \\
    \bottomrule
  \end{tabular}
  \caption{\label{tab:comparison-data}
Comparison of the pre-training effect of \datasetname~with other synthetic datasets. All datasets were trained using the Donut model.}
\end{table}

\begin{table}[t]
  \centering{
  \begin{tabular}{llll}\toprule[0.8pt]
    \textbf{Model}   & \textbf{human} & \textbf{aug.} & \textbf{avg}\\
    \midrule
    VisionTaPas~\cite{chartqa}      &   29.60      &   61.44     &      45.52               \\
    T5~\cite{t5}    &     25.12     &    56.96     &      41.04            \\
    VL-T5 ~\cite{vlt5}     &   26.24     &   56.88    &      41.56             \\
    \midrule
    Donut~\cite{donut} &      31.28  & 77.76   &   54.42                \\
  Donut+\datasetname(Ours)   &    \textbf{39.20}      & \textbf{81.20}   &   \textbf{60.84}   \\
  \midrule
    Pix2Struct~\cite{pix2struct} & 35.92    &     85.92    &    60.92       \\
    Pix2Struct+\datasetname(Ours)             & \textbf{41.84}  & \textbf{87.20} & \textbf{64.52}\\
    \bottomrule
  \end{tabular}
  }
  \caption{\label{tab:comparison-model}
Comparison of the model pre-trained on our \datasetname~to other models. 
  }
\end{table}

\begin{figure}[t]
  \centering
  \includegraphics[width=0.48\textwidth ]{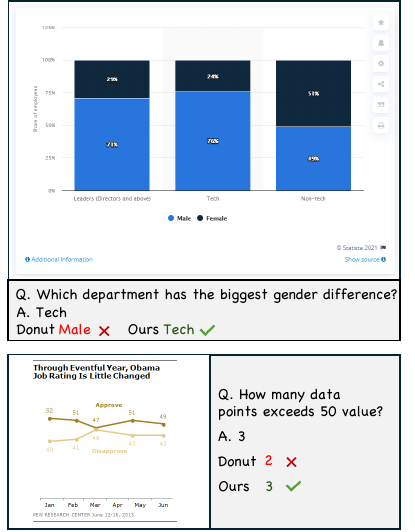}
\caption{\textbf{Qualitative Comparison.} Our pre-trained Donut model on \datasetname~demonstrates its ability to answer complex reasoning questions.
Incorrect answers are highlighted in red, while correct answers are highlighted in green. }
  \label{fig:quality}
\end{figure}

\begin{table*}[t]
\begin{tabular}{ccc}
  \begin{minipage}[t]{0.26\textwidth}
    \centering
    \scalebox{1.0}{
          \begin{tabular}{c|cc}\toprule[0.8pt]
          Randomize &  & \checkmark\\
         \midrule
           human & 33.44 & \textbf{35.92}\\
          aug. & 80.16 & \textbf{80.48}\\
        \bottomrule[0.8pt]
      \end{tabular}
    }
    \caption{\label{tab:f1}(F1) Appearance.}
  \end{minipage}

  \begin{minipage}[t]{0.26\textwidth}
    \centering
    \scalebox{1.0}{
          \begin{tabular}{c|cc}\toprule[0.8pt]
           & JSON & QA\\
         \midrule
          human  & 31.44 & \textbf{35.92} \\
          aug.  & 79.12 & \textbf{80.48}\\
        \bottomrule[0.8pt]
      \end{tabular}
    }
    \caption{\label{tab:f2}(F2) Pre-training task.}
  \end{minipage}
  \hspace{1pt}
  
  \begin{minipage}[t]{0.4\textwidth}
    \centering
    \scalebox{1.0}{
          \begin{tabular}{c|cccc}\toprule[0.8pt]
           & Template & Gemma & GPT-3.5 & GPT-4o \\
         \midrule
        human  &30.00 &31.52 &\textbf{35.92}&34.56\\
          aug. &77.92 &79.04 & 80.48&\textbf{81.84}\\
        \bottomrule[0.8pt]
      \end{tabular}
    }
    \caption{\label{tab:f3}(F3) QA quality.}
  \end{minipage}
     \end{tabular}
\end{table*}

\begin{table*}[t]
    \begin{tabular}{cccc}
    \begin{minipage}[t]{0.57\textwidth}
    \centering
    \scalebox{1.0}{
      \begin{tabular}{c|ccc}\toprule[0.8pt]
           &\textless chartqa\textgreater  & \textless synthetic\_qa\textgreater &\textless chartqa\textgreater 	\textless synthetic\_qa\textgreater \\
         \midrule 
        human&\textbf{35.92}&34.72  &33.12\\
        aug. &\textbf{80.48} &79.20&79.68\\
        \bottomrule[0.8pt]
      \end{tabular}
    }
    \caption{\label{tab:f4}(F4) Prompt.}
    \vspace{-10pt}
  \end{minipage}
  \begin{minipage}[t]{0.42\textwidth}
  \centering
   \scalebox{1.0}{
  \begin{tabular}{c|cccc}\toprule[0.8pt]
    \#Images    & 50k & 500k &  1M\\
    \midrule 
    human      &    35.92 & 36.48  & \textbf{39.20}            \\
    aug.   &   80.48 & 81.12& \textbf{81.20}   \\
        \bottomrule[0.8pt]
  \end{tabular}
  }
  \caption{\label{tab:f5}(F5) Number of images.}
    \end{minipage}
   \end{tabular} 
\end{table*}
\noindent\textbf{Model Comparison} 
We compare the model trained on our \datasetname~dataset with previous SOTA models that are not pre-trained on real-world charts and have parameter sizes less than 1 billion, the same as ours. For the Donut and Pix2struct models, we show the re-implementation result.
Our \datasetname~shows the pre-training effect on both Donut and Pix2Srtruct models.
Our \datasetname improves performance on both human and augmented splits. The human split, which includes more complex questions and answers, shows a particularly notable improvement.
Chart domain QA typically relies on costly annotated datasets. Our pipeline-generated dataset improves model performance without real-world figures, demonstrating the effectiveness of synthetic data for pre-training in figure understanding tasks.
We present qualitative comparisons for ChartQA fine-tuning between the Donut model~(fine-tuned from donut-base model) and our model (the Donut model pre-trained with our \datasetname) in Figure~\ref{fig:quality}.
This shows that our pre-trained model can answer questions requiring multi-step reasoning.

\subsection{Investigation Results}\label{investigation}

We conduct ablation studies to further investigate how our proposed pipeline impacts the model pre-training. 
In this phase, we used 50k images of our \datasetname~for efficient computational resources. 
We fine-tuned both the human and augmented (aug.) splits of the ChartQA training dataset, and tested on the Chart QA test split.
Pre-training is conducted for 3 epochs, and fine-tuning for 20 epochs.
We investigate the following five factors.

\noindent\textbf{(F1) Appearance.} 
We investigate whether the variation in the figure's appearance enhances the pre-training impact. To explore this, we compare figures created by: 1) our original proposed pipeline, which introduces randomness, such as fonts, text placement, and number presence, and 2) a modified pipeline that removes the augmentation phase, producing similar-looking figures.
Both splits use the same JSON data points for a fair comparison. 
As shown in Table~\ref{tab:f1}, our proposed pipeline, which randomizes the component appearance of the figure, achieves superior results in both test subsets. 
This indicates that a diverse range of figure structures enhances the model's generality and performance during pre-training.

\noindent\textbf{(F2) Pre-training task.} 
We investigate the effectiveness of pre-training tasks by comparing QA-based pre-training with JSON parsing-based pre-training.
For JSON parsing pre-training, we use annotated JSON generated during the data creation stage, which includes all data points, axes, and color information.
Then, we train the model to perform JSON-style data extraction from the reference figure.
As shown in Table~\ref{tab:f2}, QA type pre-training outperforms JSON parsing. To answer the QA, the model needs to extract data points along with their corresponding colors and perform reasoning based on the extracted information.
This complexity is a potential reason for the superior pre-training effect.

\noindent\textbf{(F3) QA quality.} 
We investigate how the quality of question-answer (QA) pairs affects model performance during pre-training. QAs are generated using two methods: (1) a modified pipeline utilizing 26 predefined templates with various reasoning techniques, and (2, 3, 4) our original pipeline, which uses a large language model (LLM) for few-shot prompting with sample QAs. For (2), we use the Gemma-7b model~\cite{gemma}, an open-source model based on the same technology as Gemini. 
In (3), we use GPT-3.5 Turbo, the same as our standard pipeline. 
For (4), we use GPT-4o mini, one of the latest models from OpenAI.
In Table~\ref{tab:f2}, the model trained on QAs generated by LLMs outperforms the one trained on template-based QAs in both test splits. 
GPT-3.5 Turbo achieves the highest scores on the ChartQA human split, while GPT-4o mini achieves the best result on the augmented split. These results suggest that leveraging LLMs, especially more advanced models, can significantly enhance the quality of QA pairs, leading to improved model performance.

\noindent\textbf{(F4) Prompt.}
We explore the necessity of different prompts during the pre-training phase of a synthetic dataset. 
We used three types of input prompts: A: \textless chartqa\textgreater question \textless s\_answer\textgreater (same as in the fine-tuning phase), B: \textless synthetic\_qa\textgreater question \textless s\_answer\textgreater, and C: \textless chartqa\textgreater \textless synthetic\_qa\textgreater question \textless s\_answer\textgreater. 
The same figure images and QA sets were used consistently throughout the three training processes.
Table~\ref{tab:f4} shows that prompt A achieved the best results, indicating that synthetic pre-training does not need a special prompt, even for real-world chart fine-tuning.

\noindent\textbf{(F5) Number of images.} 
We examine the effectiveness of the large number of images in the pre-training phase.
Table~\ref{tab:f5} shows that increasing the number of images enhances accuracy in both test sets. 
This indicates that our proposed pipeline, which can generate a large number of figures and QAs, plays a crucial role in figure understanding pre-training.
Furthermore, the continued improvement in accuracy as the number of images increases indicates the diversity within our dataset, contributing to more robust model training.

\subsection{Explorative Study}
Here, we conduct further experiments. 
We conduct further experiments to evaluate the generality of our \datasetname~in fine-tuning for other datasets.
We also investigate whether \datasetname still shows the pre-training effect with real-world figure pre-training datasets.
We use 1M images of \datasetname~.

\noindent\textbf{Effectiveness for other datasets.} 
We further investigate the effectiveness of pre-training on our \datasetname~for fine-tuning to other datasets. 
For PlotQA, one million training QA pairs were extracted. A 5\% relaxed accuracy is reported for 10,000 examples from each of the v1 and v2 subsets of the PlotQA test set as well as the FigureQA validation set.
We use Donut base model for both dataset fine-tuning.
We conduct one epoch of fine-tuning for both datasets. Here, 'scratch' refers to starting the fine-tuning from the donut-base model.
As shown in Table~\ref{tab:other_datasets}, \datasetname~demonstrates generality for fine-tuning datasets by showing the pre-training effect for both datasets.

\noindent\textbf{Pre-training effect for figure-specific models.}
Recently, many figure-specific models have trained on figure reasoning during pre-training~\cite{UniChart, matcha}.
Here, we evaluated whether \datasetname remains effective for pre-training figure-specific models, which are trained on real-world figures. 
For this experiment, we selected UniChart QA, a model trained on datasets of real-world charts.
We conducted the training of the UniChart QA reasoning dataset in two ways: (1) training from scratch, and (2) using a model pre-trained with \datasetname. As shown in Table~\ref{tab:UniChart}, starting with \datasetname pre-training proves to be effective for training figure QA models. This suggests that pre-training with \datasetname can efficiently enhance the performance of existing figure QA models.

\section{Conclusion}
We introduced \datasetname, a synthesized dataset generated by our proposed pipeline. 
Our pipeline enables the creation of diverse topic figures with completely accurate presence data annotations and dense QA pairs without any manual annotation. 
The model pre-trained on our dataset demonstrates a high pre-training effect for real-world figure datasets, allowing for efficient training.
We make our model, dataset, and step-by-step generation pipeline's code and prompts publicly available.

\section{Limitations}
There is room for improvement by generating more than 1M images, which we haven't tried due to computational resource constraints. Additionally, further hyper-parameter tuning for pre-training on synthesized figure data could enhance performance.

\begin{table}[t]
\centering{
  \begin{tabular}{lcccc}\toprule[0.8pt]
  & \multicolumn{2}{c}{\textbf{PlotQA}}&\multicolumn{2}{c}{\textbf{FigureQA}}\\
  \cmidrule(l{2pt}r{2pt}){2-3} \cmidrule(l{2pt}r{2pt}){4-5}
    \textbf{Pre-train}   & V1 & V2 & V1 & V2 \\
    \midrule
    Scratch      &    65.40  &       26.16  &  52.04    &    52.46     \\
    \datasetname     & \textbf{73.15}   &    \textbf{42.48}  &    \textbf{84.26}   &  \textbf{83.64}         \\
    \bottomrule
  \end{tabular}
  }
  \caption{\label{tab:other_datasets}
Evaluation of the pre-training effect of \datasetname~on the PlotQA and FigureQA tasks. All pre-training and fine-tuning were conducted using the Donut model.
  }
\end{table}

\begin{table}[t]
\centering{
  \begin{tabular}{lccc}\toprule[0.8pt]
  & \multicolumn{3}{c}{\textbf{Fine-tuning steps on UniChart QA}}\\
  \cmidrule(l{2pt}r{2pt}){2-4} 
    \textbf{Pre-train}   & 10k & 30k & 50k \\
    \midrule
    Scratch      &  31.20$\mid$78.96  & 33.60$\mid$81.12  & 34.56$\mid$78.82      \\
    \datasetname     &  \textbf{40.00$\mid$82.16}   &    \textbf{40.40$\mid$82.96}  &    \textbf{41.36$\mid$83.76}   \\
    \bottomrule
  \end{tabular}
  }
  \caption{\label{tab:UniChart}
Evaluation of the pretraining effect of our \datasetname~for the UniChart reasoning training based on steps. We evaluate on ChartQA dataset (human$\mid$aug.).
  }
  \vspace{-10pt}
\end{table}

\section{Acknowledgment}
This work was supported by JST Moonshot R\&D Program, Grant Number JPMJMS2236.
\bibliography{aaai25}

\begin{thebibliography}{22}
\providecommand{\natexlab}[1]{#1}

\bibitem[{Carbune et~al.(2024)Carbune, Mansoor, Liu, Aralikatte, Baechler, Chen, and Sharma}]{reasoning}
Carbune, V.; Mansoor, H.; Liu, F.; Aralikatte, R.; Baechler, G.; Chen, J.; and Sharma, A. 2024.
\newblock Chart-based Reasoning: Transferring Capabilities from LLMs to VLMs.
\newblock arXiv:2403.12596.

\bibitem[{Chaudhry et~al.(2020)Chaudhry, Shekhar, Gupta, Maneriker, Bansal, and Joshi}]{leafqa}
Chaudhry, R.; Shekhar, S.; Gupta, U.; Maneriker, P.; Bansal, P.; and Joshi, A. 2020.
\newblock LEAF-QA: Locate, Encode \& Attend for Figure Question Answering.
\newblock In \emph{Proceedings of the IEEE/CVF Winter Conference on Applications of Computer Vision (WACV)}.

\bibitem[{Cho et~al.(2021)Cho, Lei, Tan, and Bansal}]{vlt5}
Cho, J.; Lei, J.; Tan, H.; and Bansal, M. 2021.
\newblock Unifying Vision-and-Language Tasks via Text Generation.
\newblock In \emph{Proceedings of the 38th International Conference on Machine Learning}, volume 139 of \emph{Proceedings of Machine Learning Research}, 1931--1942. PMLR.

\bibitem[{Dosovitskiy et~al.(2021)Dosovitskiy, Beyer, Kolesnikov, Weissenborn, Zhai, Unterthiner, Dehghani, Minderer, Heigold, Gelly, Uszkoreit, and Houlsby}]{ViT}
Dosovitskiy, A.; Beyer, L.; Kolesnikov, A.; Weissenborn, D.; Zhai, X.; Unterthiner, T.; Dehghani, M.; Minderer, M.; Heigold, G.; Gelly, S.; Uszkoreit, J.; and Houlsby, N. 2021.
\newblock An Image is Worth 16x16 Words: Transformers for Image Recognition at Scale.
\newblock \emph{ICLR}.

\bibitem[{Han et~al.(2023)Han, Zhang, Chen, Yang, Wang, Yu, Fu, and Zhang}]{chartllama}
Han, Y.; Zhang, C.; Chen, X.; Yang, X.; Wang, Z.; Yu, G.; Fu, B.; and Zhang, H. 2023.
\newblock ChartLlama: A Multimodal LLM for Chart Understanding and Generation.
\newblock arXiv:2311.16483.

\bibitem[{Hoque, Kavehzadeh, and Masry(2022)}]{hoque2022chart}
Hoque, E.; Kavehzadeh, P.; and Masry, A. 2022.
\newblock Chart question answering: State of the art and future directions.
\newblock In \emph{Computer Graphics Forum}, volume~41, 555--572. Wiley Online Library.

\bibitem[{Kafle et~al.(2018)Kafle, Price, Cohen, and Kanan}]{dvqa}
Kafle, K.; Price, B.; Cohen, S.; and Kanan, C. 2018.
\newblock DVQA: Understanding Data Visualizations via Question Answering.
\newblock In \emph{IEEE/CVF Conference on Computer Vision and Pattern Recognition (CVPR)}, 5648--5656.

\bibitem[{Kahou et~al.(2018)Kahou, Michalski, Atkinson, Kadar, Trischler, and Bengio}]{figureqa}
Kahou, S.~E.; Michalski, V.; Atkinson, A.; Kadar, A.; Trischler, A.; and Bengio, Y. 2018.
\newblock FigureQA: An Annotated Figure Dataset for Visual Reasoning.
\newblock arXiv:1710.07300.

\bibitem[{Kim et~al.(2022)Kim, Hong, Yim, Nam, Park, Yim, Hwang, Yun, Han, and Park}]{donut}
Kim, G.; Hong, T.; Yim, M.; Nam, J.; Park, J.; Yim, J.; Hwang, W.; Yun, S.; Han, D.; and Park, S. 2022.
\newblock OCR-Free Document Understanding Transformer.
\newblock In \emph{European Conference on Computer Vision (ECCV)}.

\bibitem[{Lee et~al.(2023)Lee, Joshi, Turc, Hu, Liu, Eisenschlos, Khandelwal, Shaw, Chang, and Toutanova}]{pix2struct}
Lee, K.; Joshi, M.; Turc, I.; Hu, H.; Liu, F.; Eisenschlos, J.; Khandelwal, U.; Shaw, P.; Chang, M.-W.; and Toutanova, K. 2023.
\newblock Pix2Struct: screenshot parsing as pretraining for visual language understanding.
\newblock In \emph{Proceedings of International Conference on Machine Learning (ICML)}.

\bibitem[{Lewis et~al.(2020)Lewis, Liu, Goyal, Ghazvininejad, Mohamed, Levy, Stoyanov, and Zettlemoyer}]{bart}
Lewis, M.; Liu, Y.; Goyal, N.; Ghazvininejad, M.; Mohamed, A.; Levy, O.; Stoyanov, V.; and Zettlemoyer, L. 2020.
\newblock {BART}: Denoising Sequence-to-Sequence Pre-training for Natural Language Generation, Translation, and Comprehension.
\newblock In \emph{Proceedings of the Association for Computational Linguistics (ACL)}, 7871--7880. Online: Association for Computational Linguistics.

\bibitem[{Li et~al.(2024)Li, Jasani, Tang, and Ghadar}]{Li_2024_CVPR}
Li, Z.; Jasani, B.; Tang, P.; and Ghadar, S. 2024.
\newblock Synthesize Step-by-Step: Tools Templates and LLMs as Data Generators for Reasoning-Based Chart VQA.
\newblock In \emph{Proceedings of the IEEE/CVF Conference on Computer Vision and Pattern Recognition (CVPR)}, 13613--13623.

\bibitem[{Liu et~al.(2023{\natexlab{a}})Liu, Eisenschlos, Piccinno, Krichene, Pang, Lee, Joshi, Chen, Collier, and Altun}]{deplot}
Liu, F.; Eisenschlos, J.; Piccinno, F.; Krichene, S.; Pang, C.; Lee, K.; Joshi, M.; Chen, W.; Collier, N.; and Altun, Y. 2023{\natexlab{a}}.
\newblock {D}e{P}lot: One-shot visual language reasoning by plot-to-table translation.
\newblock In \emph{Findings of the Association for Computational Linguistics (ACL)}, 10381--10399. Toronto, Canada: Association for Computational Linguistics.

\bibitem[{Liu et~al.(2023{\natexlab{b}})Liu, Piccinno, Krichene, Pang, Lee, Joshi, Altun, Collier, and Eisenschlos}]{matcha}
Liu, F.; Piccinno, F.; Krichene, S.; Pang, C.; Lee, K.; Joshi, M.; Altun, Y.; Collier, N.; and Eisenschlos, J. 2023{\natexlab{b}}.
\newblock {M}at{C}ha: Enhancing Visual Language Pretraining with Math Reasoning and Chart Derendering.
\newblock In \emph{Proceedings of the 61st Annual Meeting of the Association for Computational Linguistics (Volume 1: Long Papers)}, 12756--12770. Toronto, Canada: Association for Computational Linguistics.

\bibitem[{Liu et~al.(2021)Liu, Lin, Cao, Hu, Wei, Zhang, Lin, and Guo}]{swin}
Liu, Z.; Lin, Y.; Cao, Y.; Hu, H.; Wei, Y.; Zhang, Z.; Lin, S.; and Guo, B. 2021.
\newblock Swin Transformer: Hierarchical Vision Transformer using Shifted Windows.
\newblock In \emph{Proceedings of the IEEE/CVF International Conference on Computer Vision (ICCV)}.

\bibitem[{Masry et~al.(2023)Masry, Kavehzadeh, Do, Hoque, and Joty}]{UniChart}
Masry, A.; Kavehzadeh, P.; Do, X.~L.; Hoque, E.; and Joty, S. 2023.
\newblock {U}ni{C}hart: A Universal Vision-language Pretrained Model for Chart Comprehension and Reasoning.
\newblock In \emph{Proceedings of Conference on Empirical Methods in Natural Language Processing (EMNLP)}, 14662--14684. Singapore: Association for Computational Linguistics.

\bibitem[{Masry et~al.(2022)Masry, Long, Tan, Joty, and Hoque}]{chartqa}
Masry, A.; Long, D.; Tan, J.~Q.; Joty, S.; and Hoque, E. 2022.
\newblock {C}hart{QA}: A Benchmark for Question Answering about Charts with Visual and Logical Reasoning.
\newblock In \emph{Findings of the Association for Computational Linguistics (ACL)}, 2263--2279. Dublin, Ireland.

\bibitem[{Methani et~al.(2020)Methani, Ganguly, Khapra, and Kumar}]{plotqa}
Methani, N.; Ganguly, P.; Khapra, M.~M.; and Kumar, P. 2020.
\newblock PlotQA: Reasoning over Scientific Plots.
\newblock In \emph{The IEEE Winter Conference on Applications of Computer Vision (WACV)}.

\bibitem[{Raffel et~al.(2020)Raffel, Shazeer, Roberts, Lee, Narang, Matena, Zhou, Li, and Liu}]{t5}
Raffel, C.; Shazeer, N.; Roberts, A.; Lee, K.; Narang, S.; Matena, M.; Zhou, Y.; Li, W.; and Liu, P.~J. 2020.
\newblock Exploring the Limits of Transfer Learning with a Unified Text-to-Text Transformer.
\newblock \emph{Journal of Machine Learning Research}, 21(140): 1--67.

\bibitem[{Siegel et~al.(2016)Siegel, Horvitz, Levin, Divvala, and Farhadi}]{Siegel2016FigureSeerPR}
Siegel, N.; Horvitz, Z.; Levin, R.; Divvala, S.~K.; and Farhadi, A. 2016.
\newblock FigureSeer: Parsing Result-Figures in Research Papers.
\newblock In \emph{European Conference on Computer Vision}.

\bibitem[{Singh and Shekhar(2020)}]{leafqaplus}
Singh, H.; and Shekhar, S. 2020.
\newblock {STL-CQA}: Structure-based Transformers with Localization and Encoding for Chart Question Answering.
\newblock In \emph{Proceedings of the 2020 Conference on Empirical Methods in Natural Language Processing (EMNLP)}, 3275--3284. Online: Association for Computational Linguistics.

\bibitem[{Team et~al.(2024)Team, Mesnard, Hardin, Dadashi, Bhupatiraju, Pathak, Sifre, Rivière, Kale, Love, Tafti, Hussenot, Sessa, Chowdhery, Roberts, Barua, Botev, Castro-Ros, Slone, Héliou, Tacchetti, Bulanova, Paterson, Tsai, Shahriari, Lan, Choquette-Choo, Crepy, Cer, Ippolito, Reid, Buchatskaya, Ni, Noland, Yan, Tucker, Muraru, Rozhdestvenskiy, Michalewski, Tenney, Grishchenko, Austin, Keeling, Labanowski, Lespiau, Stanway, Brennan, Chen, Ferret, Chiu, Mao-Jones, Lee, Yu, Millican, Sjoesund, Lee, Dixon, Reid, Mikuła, Wirth, Sharman, Chinaev, Thain, Bachem, Chang, Wahltinez, Bailey, Michel, Yotov, Chaabouni, Comanescu, Jana, Anil, McIlroy, Liu, Mullins, Smith, Borgeaud, Girgin, Douglas, Pandya, Shakeri, De, Klimenko, Hennigan, Feinberg, Stokowiec, hui Chen, Ahmed, Gong, Warkentin, Peran, Giang, Farabet, Vinyals, Dean, Kavukcuoglu, Hassabis, Ghahramani, Eck, Barral, Pereira, Collins, Joulin, Fiedel, Senter, Andreev, and Kenealy}]{gemma}
Team, G.; Mesnard, T.; Hardin, C.; Dadashi, R.; Bhupatiraju, S.; Pathak, S.; Sifre, L.; Rivière, M.; Kale, M.~S.; Love, J.; Tafti, P.; Hussenot, L.; Sessa, P.~G.; Chowdhery, A.; Roberts, A.; Barua, A.; Botev, A.; Castro-Ros, A.; Slone, A.; Héliou, A.; Tacchetti, A.; Bulanova, A.; Paterson, A.; Tsai, B.; Shahriari, B.; Lan, C.~L.; Choquette-Choo, C.~A.; Crepy, C.; Cer, D.; Ippolito, D.; Reid, D.; Buchatskaya, E.; Ni, E.; Noland, E.; Yan, G.; Tucker, G.; Muraru, G.-C.; Rozhdestvenskiy, G.; Michalewski, H.; Tenney, I.; Grishchenko, I.; Austin, J.; Keeling, J.; Labanowski, J.; Lespiau, J.-B.; Stanway, J.; Brennan, J.; Chen, J.; Ferret, J.; Chiu, J.; Mao-Jones, J.; Lee, K.; Yu, K.; Millican, K.; Sjoesund, L.~L.; Lee, L.; Dixon, L.; Reid, M.; Mikuła, M.; Wirth, M.; Sharman, M.; Chinaev, N.; Thain, N.; Bachem, O.; Chang, O.; Wahltinez, O.; Bailey, P.; Michel, P.; Yotov, P.; Chaabouni, R.; Comanescu, R.; Jana, R.; Anil, R.; McIlroy, R.; Liu, R.; Mullins, R.; Smith, S.~L.; Borgeaud, S.; Girgin, S.; Douglas, S.;
  Pandya, S.; Shakeri, S.; De, S.; Klimenko, T.; Hennigan, T.; Feinberg, V.; Stokowiec, W.; hui Chen, Y.; Ahmed, Z.; Gong, Z.; Warkentin, T.; Peran, L.; Giang, M.; Farabet, C.; Vinyals, O.; Dean, J.; Kavukcuoglu, K.; Hassabis, D.; Ghahramani, Z.; Eck, D.; Barral, J.; Pereira, F.; Collins, E.; Joulin, A.; Fiedel, N.; Senter, E.; Andreev, A.; and Kenealy, K. 2024.
\newblock Gemma: Open Models Based on Gemini Research and Technology.
\newblock arXiv:2403.08295.

\end{thebibliography}
\appendix 
\section{Implementation details}
\label{sec:intro}
We provide the implementation details in the main paper. Here, we offer additional implementation specifics. Our work primarily utilizes the Donut model~\cite{donut}, as described in the main paper, and we also use Pix2Struct~\cite{pix2struct} to demonstrate the generality of our approach. The training details are as follows;

\begin{itemize}
    \item Donut~\cite{donut}: Following the UniChart~\cite{UniChart}, which is based on Donut, we adopt the same hyper-parameters. The batch size is set to 80 during pre-training and 24 during fine-tuning. The learning rate is 5e-5 for pre-training and 1e-4 for fine-tuning. Additionally, the image resolution is fixed at $960 \times 960$. 
    \item Pix2Struct~\cite{pix2struct}: We adopt the same hyper-parameters as UniChart~\cite{UniChart}, with two differences: the pre-training batch size is set to 40 due to the difference in model parameter size, and the image resolution is adjusted to $640 \times 640$.
\end{itemize}

In Table 2, we present the re-implementation results for Donut and Pix2Struct, allowing a fair comparison between training from scratch and using \datasetname under the same settings. For VisionTaPas~\cite{chartqa}, T5~\cite{t5}, and VL-T5~\cite{vlt5}, we report the official results from their respective publications.

\section{Dataset details}
We provide the dataset details in the main paper. 
In Table~\ref{tab:details-dataset}, we present the dataset details used for pre-training in the main paper for clearer understanding. 
The number of images in \datasetname varies across experiments
(e.g., 50k images in Tables 3–6, a variable number in Table 7, and 1M images in the other tables). We use v1 split of PlotQA~\cite{plotqa}.
\begin{table}[t]
  \centering
  \begin{tabular}{lll}\toprule[0.8pt]
    \textbf{Dataset}    & \#Images &	\#QAs\\
    \hline
    FigureQA&	180k&	2.3M                \\
DVQA	&300k	&3.4M \\
PlotQA	&224k&	8M         \\
    \datasetname(Ours)   &   50k - 1M   & 2.6k - 4.2M     \\
    \bottomrule
  \end{tabular}
  \caption{\label{tab:details-dataset}
\textbf{Comparison of the pre-training effect of \datasetname~with other synthetic datasets.} All datasets were trained using the Donut model.
  }
\end{table}
\begin{figure}[t]
  \centering
  \includegraphics[width=0.89\linewidth]{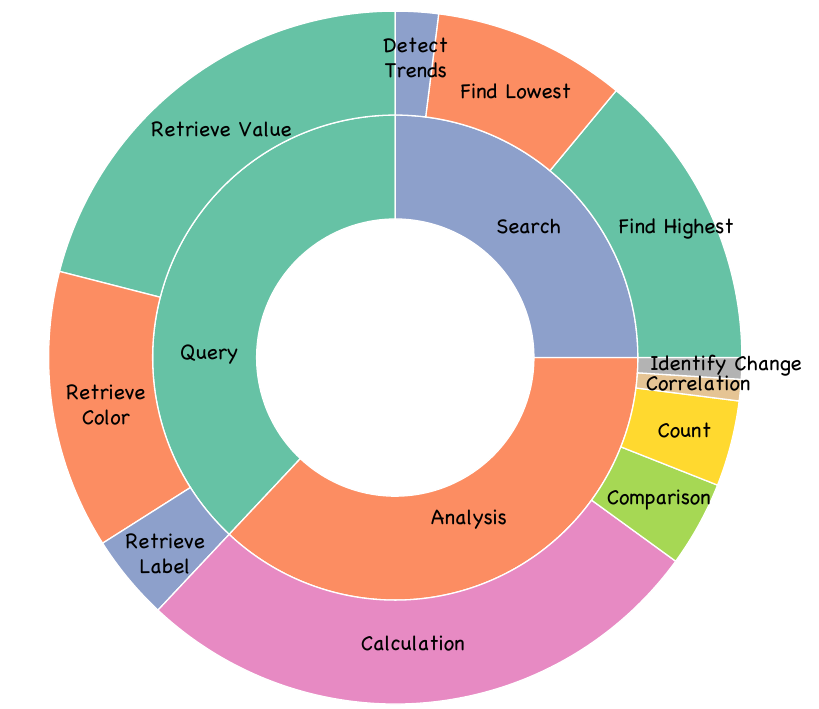}
\caption{\textbf{QA distribution of \datasetname.} We randomly selected 100 QAs and manually analyzed their QA types.}
  \label{fig:ex_qa_stat}
\end{figure}

\begin{figure*}[t]
  \centering
  \includegraphics[height=19cm]{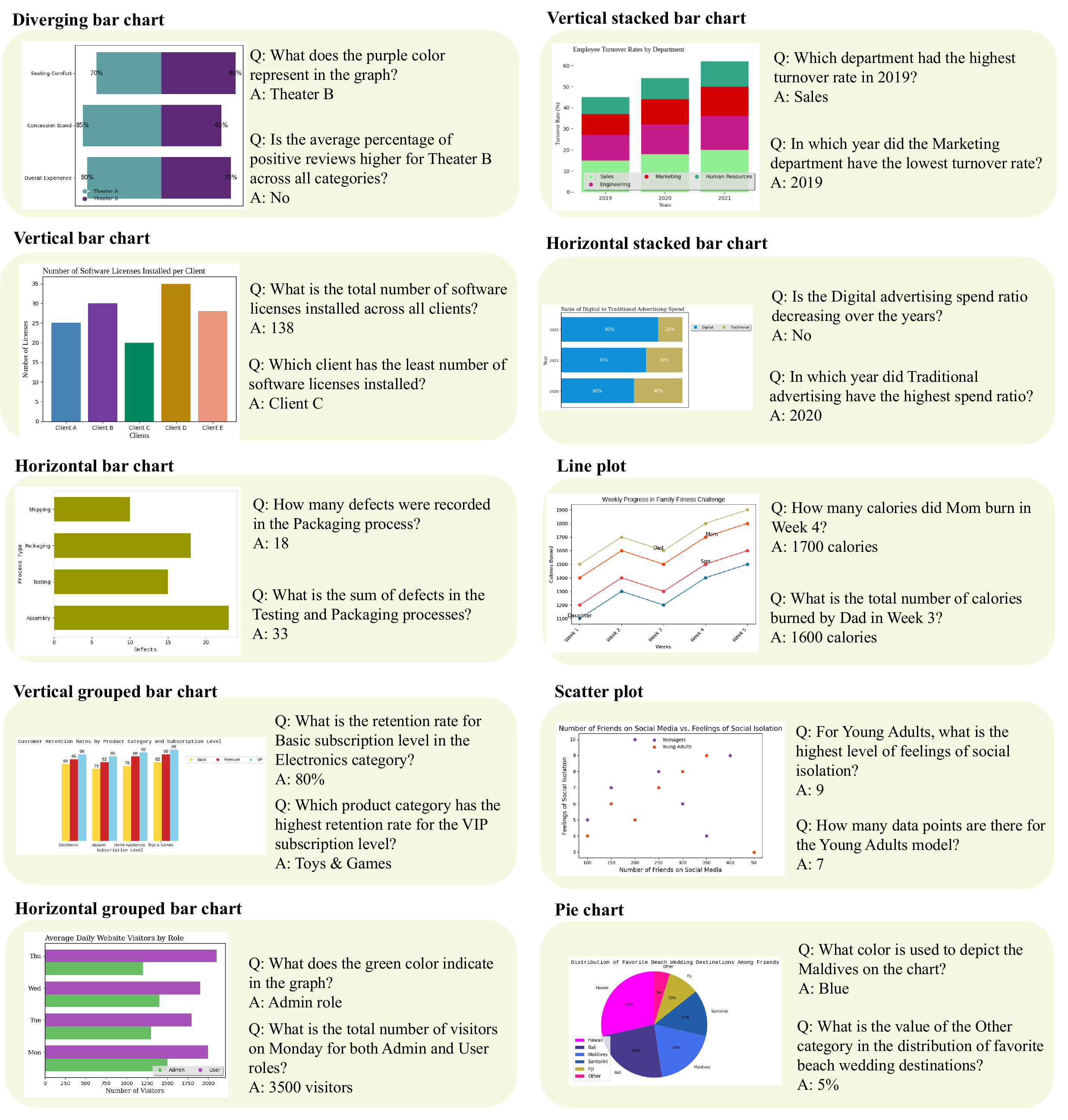}
\caption{Examples of \datasetname~figure images and QA pairs.} 
  \label{fig:ex_qa_supp}
\end{figure*}

We also present the statistics of our \datasetname QAs. A random sample of 100 QAs was extracted and manually analyzed based on their hierarchy of categories.
As illustrated in Figure~\ref{fig:ex_qa_stat}, our \datasetname effectively generates a diverse range of QAs.
Additional figure images and examples of QAs are provided in Figure~\ref{fig:ex_qa_supp}.
\end{document}